\documentclass{article}
\usepackage{spconf,amsmath,graphicx}
\usepackage{amsfonts} 
\usepackage{amsmath} 
\usepackage{graphicx} 
\usepackage{multirow} 
\usepackage{booktabs} 

\title{COARSE-TO-CAREFUL: SEEKING SEMANTIC-RELATED KNOWLEDGE FOR OPEN-DOMAIN COMMONSENSE QUESTION ANSWERING}
\name{Luxi Xing \textsuperscript{1,2}, Yue Hu\textsuperscript{1,2}\sthanks{Corresponding author. E-mail: huyue@iie.ac.cn}, Jing Yu\textsuperscript{1,2}, Yuqiang Xie\textsuperscript{1,2}, Wei Peng\textsuperscript{1,2}}
\address{\textsuperscript{1}Institute of Information Engineering, Chinese Academy of Sciences, China \\
\textsuperscript{2}School of Cyber Security, University of Chinese Academy of Sciences, China}

\begin{document}
%
\maketitle
\begin{abstract}
It is prevalent to utilize external knowledge to help machine answer questions that need background commonsense, which faces a problem that unlimited knowledge will transmit noisy and misleading information.
Towards the issue of introducing related knowledge, we propose a semantic-driven knowledge-aware QA framework, which controls the knowledge injection in a coarse-to-careful fashion.
We devise a tailoring strategy to filter extracted knowledge under monitoring of the coarse semantic of question on the knowledge extraction stage.
And we develop a semantic-aware knowledge fetching module that engages structural knowledge information and fuses proper knowledge according to the careful semantic of questions in a hierarchical way.
Experiments demonstrate that the proposed approach promotes the performance on the CommonsenseQA dataset comparing with strong baselines.
\end{abstract}
\begin{keywords}
Machine Reading Comprehension, Commonsense Knowledge
\end{keywords}
\section{Introduction}
\label{sec:intro}

Open-domain CommonSense Question Answering (OpenCSQA) is a challenging research topic in AI,
which aims at evaluating that if a machine can give the correct answer with the ability to manipulate knowledge like human beings.
Questions in OpenCSQA require to be answered with the support of rich background commonsense knowledge.
Though it is trivial for human to solve such questions merely based on the question, it is awkward for machines to figure out.


It is notable that emerging Pre-trained Language Models (PLMs), such as BERT \cite{devlin-etal-2019-bert}, can achieve remarkable success on a variety of QA tasks.
The outstanding performance of PLMs benefits from the large scale textual corpus \cite{devlin-etal-2019-bert,2019-roberta}.
However, commonsense knowledge is often not accessible in the textual form \cite{DBLP:journals/cacm/DavisM15}.
PLMs also struggle to answer questions that require commonsense knowledge beyond the given textual information from the question.
Specifically, they make the judgment with their encapsulated knowledge in unconsciousness manner and emphasize the textual matching between words in the question and answer \cite{DBLP:journals/corr/abs-1910-12391}.

There have been several attempts on OpenCSQA regarding leveraging external structured knowledge.
Some only incorporate the entity \cite{yang-etal-2019-ktnet} or the relation between words \cite{wang-etal-2018-yuanfudao, wang-jiang-2019-kar} from the knowledge base while ignoring the structural information.
Due to the absence of the background document in the OpenCSQA task, it is hard to restrain the boundary of the knowledge to be grounded.
Most works search and rank the selected knowledge triples based on the heuristic occurrence score \cite{mihaylov-frank-2018-knowledgeable, bauer-etal-2018-commonsense} or the score function of the knowledge representation learning \cite{lin-etal-2019-kagnet}.
Also, previous works pay no attention to the contextual semantic relevance of the extracted knowledge with the question.
Massive external knowledge will bring in noisy and misleading information, and consequently, lead to deviating out of the content of the current question.
Therefore, how to introduce knowledge semantically related to questions remains substantially challenging.

To address this issue, we propose a 
\textbf{\underline{Se}mantic-driv\underline{e}n \underline{K}nowledge-aware} \textbf{Q}uestion \textbf{A}nswering 
(\textbf{SEEK-QA}) 
framework to manipulate the injection of the relevant knowledge triples in a \emph{coarse-to-careful} fashion.
The advancement of the proposed \textbf{SEEK-QA} framework lies in two folds.
First,
we design a 
\textbf{Semantic M\underline{o}nitoring K\underline{n}owledge T\underline{a}ilo\underline{r}ing}
(\textbf{SONAR}) strategy
which can constrain the selected knowledge triples with the global \emph{coarse} semantic of the question.
It not only denoises irrelevant knowledge input but also benefits upgrading the computing efficiency of the follow-up model.
Second,
we develop
\textbf{\underline{S}emantic-aware \underline{K}nowledge F\underline{etch}ing}
(\textbf{SKETCH}) module, 
which capitalizes the \emph{careful} semantic of question to measure the semantic relevance between the question and knowledge triplets in a hierarchical way.

In this paper, we focus on the typical multiple-choice question answering benchmark CommonsenseQA \cite{talmor-etal-2019-commonsenseqa} and consider the structured knowledge base {ConceptNet} \cite{2017-conceptnet} as the external  knowledge source.
Experiment results demonstrate that \textbf{SONAR} and \textbf{SKETCH} can boost the performance compared with strong baselines.
And we exhibit that how injected knowledge influences the judgment of model.

\section{Overview of SEEK-QA}
\label{sec:overview}


In the CommonsenseQA task, given a question $\mathbf{Q}$ and a set of candidate answers $\mathcal{A}$, the model $\mathcal{M}$ is asked to select the only one correct answer $\mathbf{A}^{\ast}$ from the candidate set.
When involving the external knowledge $\mathcal{K}$, the goal of the model can be formalize as:
$\mathbf{A}^{\ast} = \mathop{\arg\max}_{A_i \in \mathcal{A}} \mathcal{M}(A_i| \mathbf{Q}, \mathcal{K}_i)$,
where $\mathcal{K}_i$ stands for requisite knowledge extracted from {ConceptNet} of the $i$-th candidate answer.
Each knowledge in {ConceptNet} is indicated in a triple $k = ({cn}^{head}, r, {cn}^{tail})$, where $cn^{head}$, $cn^{tail}$ are concepts and $r$ is relation between them.
We denote extracted knowledge graph of $\mathcal{K}_i$ as $\mathcal{G}=\{\mathcal{V}, \mathcal{E} \}$, where $\mathcal{V}$ represents concepts, and $\mathcal{E}$ stands for relations.

\begin{figure}[t!]
  \centering
  \includegraphics[scale=0.07]{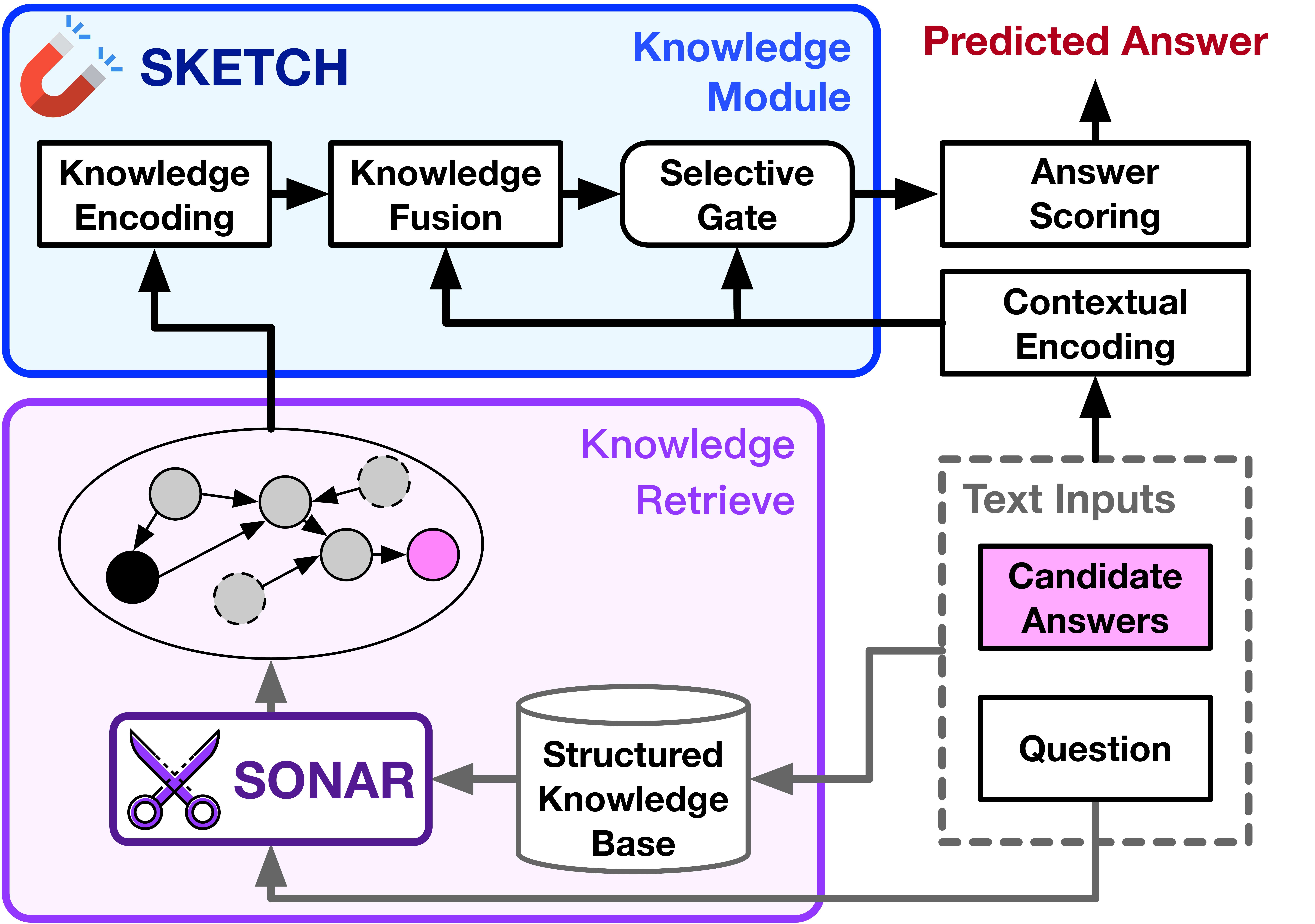}
  \caption{Workflow of Semantic-driven Knowledge-aware Question Answering (SEEK-QA) framework.}
  \label{fig:seek-qa}
\end{figure}

Our \textbf{SEEK-QA} framework, as shown in Fig.\ref{fig:seek-qa}, follows the \emph{retrieve-then-answering} paradigm that contains two phases: 
a) \emph{retrieve}: using the \textbf{SONAR} strategy (Section \ref{sec:kn-tailor}) to filter irrelevant knowledge triples.
b) \emph{answering}: utilizing a QA model equipped with \textbf{SKETCH} module to select the correct answer for a given question.
%

The main QA model consists of three stages: 

\textbf{Contextual Encoding:}
We utilize a PLM as the contextual encoder.
It takes question $\mathbf{Q}$ and each candidate answer $A_i$ ($[\texttt{CLS}]\mathbf{Q}[\texttt{SEP}]A_i[\texttt{SEP}]$) as input.
The contextual representation of the $t$-th token is denoted as $h_t^c \in \mathbb{R}^{d_h}$, and $d_h$ is the hidden dimension of the PLMs encoder.
We treat the output of \texttt{[CLS]} ($h_0^c \in \mathbb{R}^{d_h}$) as the global semantic representation of the question-answer pair.


\textbf{Knowledge Module:}
The knowledge module, {SKETCH}, is the core of QA model and will be elaborated in Section \ref{sec:kn}.
It is responsible for encoding and fusing knowledge and generating integrated information $I$ for making final prediction.

\textbf{Answer Scoring:}
The last step is to calculate the final score for each candidate answer as:
$\mbox{score}(A_i | \mathbf{Q}, \mathcal{K}_i) = \mbox{MLP}(I)$,
where $I$ will be presented in Section \ref{subsec:kn-fusion}.
The final probability for the candidate answer $A_i$ to be selected can be formulated as:
$\mathbf{P}(A_i|\mathbf{Q},\mathcal{K}_i) = \frac{\exp\left(\mbox{score}(A_i)\right)}{\sum_{i^{\prime}=1}^{\left\vert\mathcal{A}\right\vert} \exp\left( \mbox{score}(A_{i^{\prime}}) \right)}$.

Here, we use cross entropy to calculate the losses between predictions of the model and the ground-truth answer labels.

\section{Semantic Monitoring Knowledge Tailoring}
\label{sec:kn-tailor}
In this section, we present the \textbf{SONAR} strategy for knowledge selecting and filtering.
During knowledge selection,
we first convert question and answer into a list of concepts separately, indicated as $\mathcal{C}n^q$ and $\mathcal{C}n^a$.
We denote the $m$-th concept in question as $cn^q_m \in \mathcal{C}n^q$ and $n$-th concept in answer as $cn^a_n \in \mathcal{C}n^a$.
Next, we extract the link paths between question concept and answer concept with the maximal hop to $3$.
In the meanwhile, we extract link paths between two arbitrary question concepts.
After the knowledge selection, we obtain multiple groups of link paths between each pair of concepts $\langle cn^q_m, cn^a_n \rangle$.

During the knowledge filtration, 
we first adopt the knowledge embedding algorithm, i.e., TransE \cite{2013-nips-transe}, on ConceptNet to obtain the knowledge embedding of concepts and relations.
Next, the SONAR rates link paths from three aspects:
(1) On the link aspect, we follow operation in \cite{lin-etal-2019-kagnet} that defines the score of one link path as the product of validity score of each triple.
(2) On the concept aspect, SONAR represents question with GloVe embedding
and operates mean pooling over sentence length to get the question representation.
Meanwhile, the concept is represented with its GloVe embedding.
{SONAR} strategy utilizes cosine similarity to measure the semantic relevance score between the question representation and each concept representation, and the final link path score is defined as the average of all concepts' scores.
(3) On the relation aspect, SONAR performs the same operation as the concepts, while the relation is represented with its embedding acquired from the knowledge embedding algorithm.
Finally, SONAR removes the link path that does not meet up the bond of thresholds on all three scores and preserves the link path satisfying at least two thresholds.
After filtering, the reserved link paths are gathered into a knowledge sub-graph $\mathcal{G}$ that will be taken as the input for the follow-up QA model.

\section{Semantic-aware Knowledge Fetching}
\label{sec:kn}
In this section, we elaborate on the workflow of the knowledge module, \textbf{SKETCH}, over the acquired knowledge.

\subsection{Knoweledge Encoding}
\label{subsec:kn-enc}
Considering that structural relations offer positive semantic information, 
we employ the relation-aware graph attention network \cite{DBLP:journals/corr/abs-1710-10903, DBLP:conf/ijcai/ZhouYHZXZ18} to calculate graph level representation as :
\begin{eqnarray}
  \label{eq:rgat}
  && \boldsymbol{h}_j^{g} = \sum\limits_{j^{\prime}=1}^{|{Nb}_j|} \boldsymbol{\alpha}_{ j^{\prime} } [\boldsymbol{\hat{h}}_j^{g}; \boldsymbol{\hat{h}}_{ j^{\prime} }^g] , \boldsymbol{\alpha}_j = \frac{\mbox{exp}(\boldsymbol{\beta}_j)}{\sum_{j^{\prime}=1}^{|{Nb}_j|} \mbox{exp}(\boldsymbol{\beta}_{j^{\prime}})} \\
  && \boldsymbol{\beta}_j = (W_r \boldsymbol{r}_j)^{\top} \mbox{tanh} (W_{1} \boldsymbol{\hat{h}}_j^{g} + W_{2} \boldsymbol{\hat{h}}_{j^{\prime}}^g)
\end{eqnarray}
where $Nb_j$ is the set of neighbor nodes of $j$-th node, 
$\boldsymbol{\hat{h}}_{*}^{g}$ and $\boldsymbol{h}_{*}^g \in \mathbb{R}^{d_g}$ are the concept representations,
$\boldsymbol{r}_j \in \mathbb{R}^{d_r}$ is the relation representation,
and $W_r$, $W_1$, $W_2$ are trainable matrices.

Each link path $l_k$ is a sequence of concatenation triples as: 
$cn_1 \stackrel{r_1}{\longrightarrow} cn_2 \dots\dots \xrightarrow{r_{|l_k| - 1}} cn_{|l_k|}$.
We employ a heuristic method to encode the $k$-th link path with bidirectional GRU \cite{cho-etal-2014-learning}:
$h_{k_j}^l = \mbox{BiGRU}([{h}_j^g; r_j; {h}_{j+1}^g])$,
Then, we compute a single vector of the link path representation through mean pooling over its sequential representation: $u_k = \mbox{mean}(h_k^l)$, which is taken as the knowledge representation of $k$-th link path.

\subsection{Knowledge Fusion}
\label{subsec:kn-fusion}

\begin{figure}[t!]
  \centering
  \includegraphics[scale=0.065]{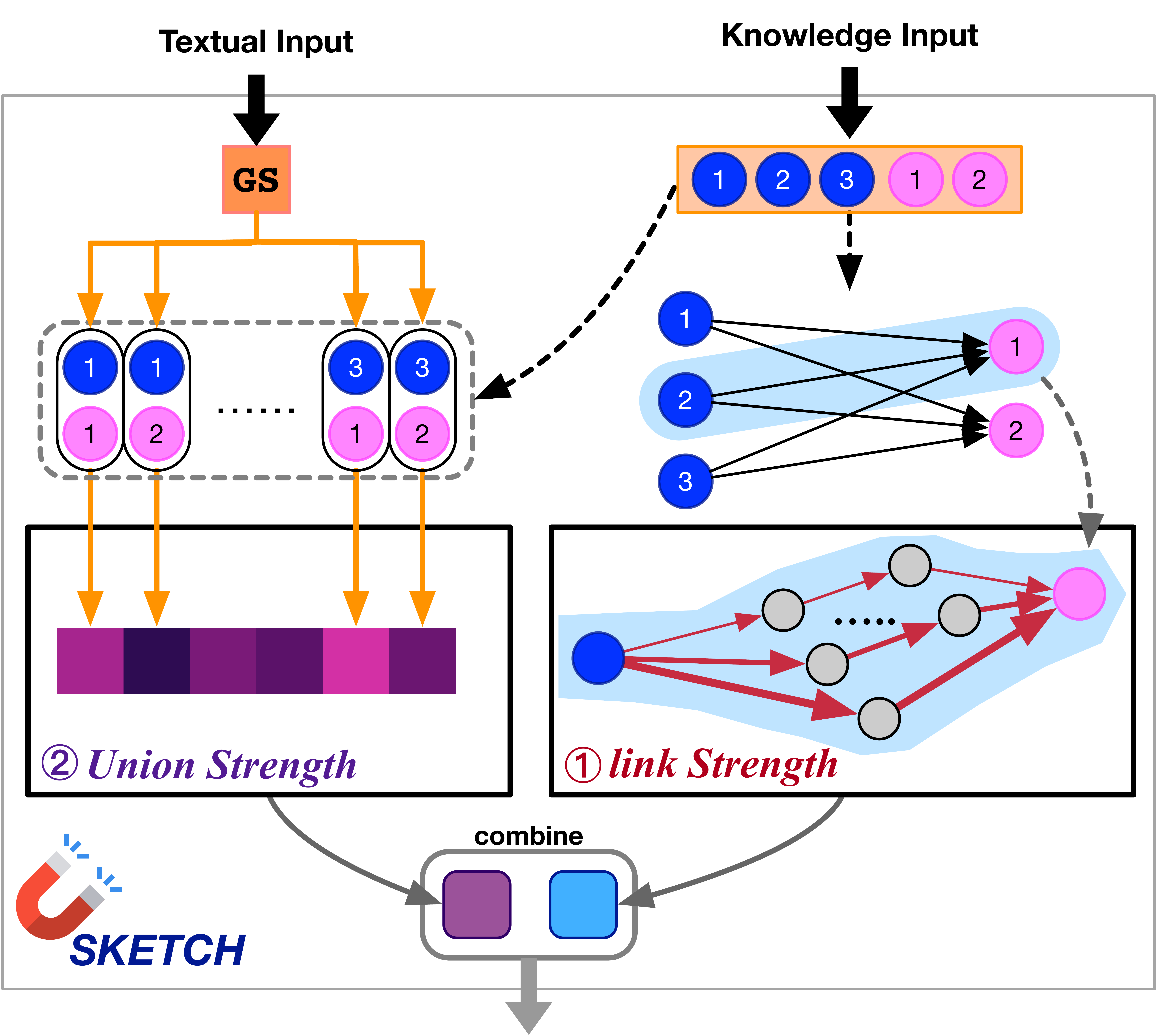}
  \caption{Progressive knowledge fusion mechanism inside the SKETCH.
  \texttt{GS}: global semantic representation.
  Blue circles represent concepts in question, while pinks correspond to concepts in answer.
  Arrow lines in red indicate \emph{link strength}.
  The thicker the line, the higher the strength.
  Purple rectangles indicate \emph{union strength}.
  The darker, the higher the strength.}
  \label{fig:sketch-strength}
\end{figure}

During fusion phase, SKETCH is equipped with a semantic-aware progressive knowledge fusion mechanism, as shown in Fig. \ref{fig:sketch-strength}, to integrate relevant knowledge considering that different pairs of concepts dedicate diverse impact to question.

We treat the link paths containing the same $k^{\prime}$-th concept pair $\langle cn^q_m, cn^a_n \rangle$ as a link group $O_{k^{\prime}}$.
For a group of link paths, we calculate the \emph{semantic link strength} which implies the semantic relevance between a link path $l_k \in O_{k^{\prime}}$ and concepts pair $\langle cn^q_m, cn^a_n \rangle$ 
as:
$\boldsymbol{\alpha}_k^{l} = (W_3[h_m^c; h_n^c])^{\top} (W_4 u_k)$,
where $h_m^c$ is representation from contextual encoder output of $cn_m^q$ while $h_n^c$ is for $cn_n^a$, and $W_3$, $W_4$ are trainable matrices.
Then \emph{semantic link strength} $\boldsymbol{\alpha}^{l}$ is normalized within its group and assemble the representation of link group $O_{k^{\prime}}$ as follows:
\begin{equation}
    U_{k^{\prime}} = \sum_{k=1}^{\left\vert O_{k^{\prime}} \right\vert} \boldsymbol{\beta}_k^l u_k , \boldsymbol{\beta}_k^{l} = \frac{ \mbox{exp}(\boldsymbol{\alpha}_k^l) }{ \sum_{l_s \in O_{k^{\prime}}} \mbox{exp} (\boldsymbol{\alpha}_s^l) }
\end{equation}

Among different pairs of concepts, \emph{semantic union strength} is designed to fetch semantic relevance between a pair of concepts and the global question semantic which expounds how well the concept pair contributes to the question.
The \emph{semantic union strength} is calculated as follows:
\begin{equation}
  \boldsymbol{\alpha}_{k^{\prime}}^{c} = (W_5 h_0^c)^{\top} (W_6 [h^g_{m}; h^g_{n}; h_m^c; h_n^c])
\end{equation}
where $h^g_{m}$ is the graph-level representation of $cn_m^q$,
$h^g_{n}$ is the graph-level representation of $cn_n^a$,
and $W_5$, $W_6$ are trainable weight matrics.
Then the \emph{semantic union strength} $\boldsymbol{\alpha}^{c}$ is normalized as:
$\boldsymbol{\beta}_{k^{\prime}}^{c} =  \exp(\boldsymbol{\alpha}_{k^{\prime}}^c) / \sum_{ s^{\prime} = 1}^{\left\vert \langle \mathcal{C}n^q, \mathcal{C}n^a \rangle \right\vert} \exp (\boldsymbol{\alpha}_{k^{\prime}}^c) $.


Combining the \emph{semantic union strength} and \emph{semantic link strength}, we can obtain the final semantic-aware knowledge representation as:
$V^k = \sum_{ k^{\prime} = 1 }^{\left\vert \langle \mathcal{C}n^q, \mathcal{C}n^a \rangle \right\vert} \boldsymbol{\beta}_{k^{\prime}}^{c} \cdot F \left( [ h^g_{m}; h^g_{n} ; U_{k^{\prime}} ] \right)$,
where $V^k \in \mathbb{R}^{d_k}$,
$F(\cdot)$ is $1$-layer feed-forward network.

For candidate answer $A_i$, we extract its graph-level representation $h_n^g$ as knowledgeable representation $V^a$.
If one candidate answer contains more than one concept, we calculate mean pooling of graph-level representation of concepts.

In the last,
SKETCH employs a selective gate, which gathers semantic-aware knowledge representation, graph-level knowledgeable representation of candidate answer, global question semantic representation together, to construct the final output for the answer scoring module as follows:
\begin{eqnarray}
  && V = F( [V^k; V^a] ) , z = \sigma( W_z [h_0^c; V] ) \\
  && I = z \cdot F( h_0^c ) + (1-z) \cdot V
\end{eqnarray}
where $W_z$ is a trainable matrix.
The $z$ controls selective merging information from external knowledge.

\section{Experiments}
\label{sec:exps}

\subsection{Dataset and Knowledge Source}
We examine our proposed approach on CommonsenseQA dataset \cite{talmor-etal-2019-commonsenseqa}, a multiple choices question answering dataset.
Each example consists of $1$ question and $5$ candidate answers.
Because of the test set is unpublicized, we randomly split official development set into one inner development set ($60\%$) and one inner test set ($40\%$) for conducting model selection and ablation studies.
The results are evaluated with accuracy.
We consider ConceptNet as our external knowledge source.
After removing non-English concepts, 
there are $2,487,809$ triples with $799,272$ concepts and $40$ relations.
These triples are taken as the input to the knowledge extraction (SONAR).
The thresholds in SONAR for link/concept/ relation are set to $0.15$/$0.3$/$0.35$
and max hop is set to $2$.
Refer to Table \ref{tb:cn-data} for the detailed statistics of obtained link paths.

\begin{table}[htb!]
  \centering
  \resizebox{0.28\textwidth}{!}{ 
  \begin{tabular}{l|cccc}
  \hline\hline
  \textbf{Num.} & \textbf{\begin{tabular}[c]{@{}c@{}}Orig. 3\end{tabular}} & \textbf{\begin{tabular}[c]{@{}c@{}}Hop 3\end{tabular}} & \textbf{\begin{tabular}[c]{@{}c@{}}Hop 2\end{tabular}} & \textbf{\begin{tabular}[c]{@{}c@{}}Hop 1\end{tabular}} \\ \hline\hline
  {\textit{Total} L}       & 41148k          & 15634k          & 1547k           & 52k                \\ \hline
  {\textit{Avg.} L1}       & 845             & 321             & 32              & 1                  \\ \hline
  {\textit{Avg.} CP}       & 10              & 10              & 10              & 10                 \\ \hline
  {\textit{Avg.} L2}       & 84              & 31              & 3               & \textless{}1       \\ \hline\hline
  \end{tabular}
  }
  \caption{Statistics of the extracted link paths with SONAR with different max hops.
  L: link paths.
  CP: numbers of concept pairs of one QA pair.
  L1/L2: numbers of links of one QA pair/pair of concepts.
  Orig.$3$: link paths without filtering.}
  \label{tb:cn-data}
\end{table}

\subsection{Experiment Setups}
We implement our approach with Pytorch \cite{pytorch:2017} framework.
We employ RoBERTa large \cite{2019-roberta} as  contextual encoder.
Dimension of embeddings of concepts and relations is set to $100$.
The layers of graph attention are $2$ with the both hidden size of $100$.
The hidden size of one-layer BiGRU for link path encoding is $150$.
Max length of textual input is $100$.
We adopt Adam \cite{DBLP:journals/corr/KingmaB14} optimizer with initial learning rate of $0.00001$.
We train our model for $1200$ steps with a batch size of $24$.
\begin{table}[htb!]
  \centering
  \resizebox{0.24\textwidth}{!}{
  \begin{tabular}{c|l|c}
  \toprule
  \textbf{Group}     & \textbf{Model}          & \textbf{Acc. (\%)}     \\ \midrule
  \multirow{3}{*}{1} & BERT-large              & 63.64                  \\
                     & BERT-wwm                & 65.40                  \\
                     & RoBERTa-large           & 71.44                  \\ \midrule
  2                  & KagNet \cite{lin-etal-2019-kagnet}   & 64.46     \\ \midrule
  ours               & \textbf{SEEK-QA}        & \textbf{74.52}         \\
  \bottomrule
  \end{tabular}
  }
  \caption{The performance on CommonsenseQA.}
  \label{tb:result}
\end{table}
\subsection{Results}

The main results of baseline models and ours on CommonsenseQA are shown in Table \ref{tb:result}.
We compare our model with two different groups of models on the CommonsenseQA task.
\textbf{Group $1$:} models with {inner encapsulated knowledge}, which including the PLMs, such as GPT \cite{2018_gpt} and BERT \cite{devlin-etal-2019-bert}.
\textbf{Group $2$:} models with {explicit structured knowledge} following the retrieve-then-answering paradigm.
Our SEEK-QA achieves a promising improvement over baselines.
For Group $1$,
it is obvious that our approach promotes the performance of PLMs through introducing appropriate knowledge.
For Group $2$, our approach also exhibits its advantage from the results.

\section{Analysis and Discussion}
\label{sec:analysis}


\begin{table}[t]
  \centering
  \resizebox{0.25\textwidth}{!}{
  \begin{tabular}{l|ccc}
  \toprule
  \textbf{Acc. (\%)}         & \textbf{Hop=3}       & \textbf{Hop=2}       & \textbf{Hop=1}       \\ \midrule
  \textbf{SONAR}             & \textbf{73.21}       & \textbf{74.52}       & \underline{75.10}       \\ 
  \textbf{w/o SC}            & 72.48                & 72.64                & 73.28                \\ 
  \textbf{w/o filter}        & --                   & {73.62}    & \textbf{75.14}       \\ \bottomrule
  \end{tabular}
  }
  \caption{Ablation studies on knowledge range on dev set.
  SC: semantic constraint scores of concept/relation in Section \ref{sec:kn-tailor}.}
  \label{tb:ablation-exp-kn}
\end{table}
\begin{table}[t]
  \centering
  \resizebox{0.25\textwidth}{!}{
  \begin{tabular}{l|cc|cc}
    \toprule
    \multirow{2}{*}{\textbf{Acc. (\%)}} & \multicolumn{2}{c|}{\textbf{Hop=2}} & \multicolumn{2}{c}{\textbf{Hop=1}} \\ \cmidrule{2-5} 
                                           & \textbf{sDev}    & \textbf{sTest}   & \textbf{sDev}   & \textbf{sTest}   \\ \midrule
    \multicolumn{1}{l|}{\textbf{SKETCH}}   & \textbf{74.04}   & \textbf{76.27}   & \textbf{74.59}  & \textbf{75.86 }  \\
    \multicolumn{1}{l|}{\textbf{GATL=1}}   & 70.90            & 74.84            & 73.90           & 75.01            \\
    \multicolumn{1}{l|}{\textbf{GATL=3}}   & 72.95            & 75.25            & 71.72           & 72.59            \\
    \multicolumn{1}{l|}{\textbf{w/o GAT}}  & 70.49            & 71.98            & 72.54           & 73.41            \\
    \multicolumn{1}{l|}{\textbf{w/o SLS}}  & 72.40            & 75.25            & 71.17           & 73.41            \\
    \multicolumn{1}{l|}{\textbf{w/o SUS}}  & 70.49            & 73.00            & 71.72           & 73.64            \\ \bottomrule
  \end{tabular}
  }
  \caption{Ablation studies on SKETCH model. 
  GATL: Layers of Graph Attention network (GAT).
  SLS: Semantic Link Strength.
  SUS: Semantic Union Strength.}
  \label{tb:ablation-exp-module}
\end{table}

\begin{figure}[t!]
  \centering
  \includegraphics[scale=0.07]{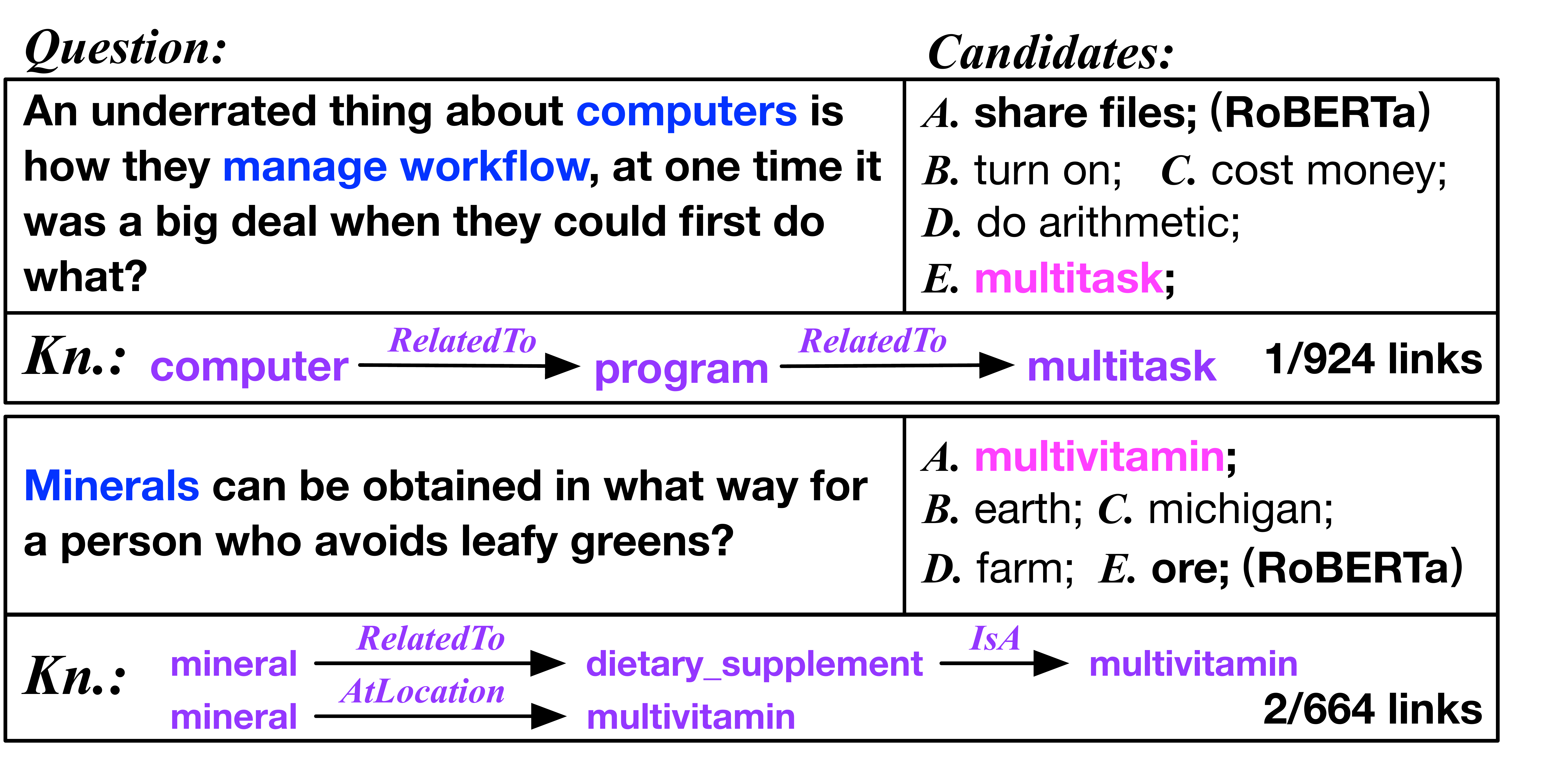}
  \caption{Case study. 
  \textit{Kn.} contains selected knowledge through SONAR and held/original links amount of QA pair.
  The correct answer is in bold with pink.}
  \label{fig:case-study}
\end{figure}

\textbf{(1) Ablation on the Range of External Knoweledge:}
We compare the impact of ranges of introduced knowledge extracting with SONAR strategy or filtering knowledge without contextual semantic constraints.
As shown in Table \ref{tb:ablation-exp-kn}, 
filtering knowledge triplets with SONAR results in a better performance comparing with removing semantic constraint.
We argue that decreasing the input of irrelevant noisy knowledge can make QA model more focused.
Comparing with removing filtering\footnote{i.e. All extracted knowledge is taken as the input to QA model.}, SONAR strategy also shows an advantage on a wider range of knowledge (Hop=$2$,$1$).
That reveals SONAR strategy can help to attract more suitable knowledge for the task and be of benefit to the task performance.
\textbf{(2) Ablation on SKETCH Component:}
As shown in Table \ref{tb:ablation-exp-module},
the performance first increases then decays along with the increasing of GATL, and drops when we remove the GAT knowledge encoding.
We can assume that GAT can facilitate model to gather information from structured knowledge.
When SLS and SUS operations are removed respectively, the performance both gets a decline.
It indicates semantic relevance strength helps to distinguish the worth of knowledge triplets.
\textbf{(3) Case Study}:
As shown in Fig.\ref{fig:case-study}, RoBERTa fails in these cases, but our approach makes a correct prediction with the support of the closely related knowledge links.
It is notable that the amount of links is greatly reduced with the SONAR strategy while core knowledge links are held.
With such requisite knowledge taken as input, the SKETCH model figure out correct answers with great confidence.

\section{Conclusion}
In this work, we propose the Semantic-driven Knowledge-aware Question Answering ({SEEK-QA}) framework, which can manipulate the injection of external structured knowledge in the light of a coarse-to-careful fashion.
Experiment results demonstrate the effectiveness of the proposed approach.

\section*{Acknowledgments}
We thank the reviewers for their advice.
This work was supported by in part by the National Key Research and Development Program of China under Grant No.2016YFB0801003.



\bibliographystyle{IEEEbib}
\bibliography{refs-xlx}

\end{document}